\documentclass{WileyMSP-template}

\usepackage{bibentry}
\usepackage{enumitem} 
\usepackage{times}
\usepackage{caption}
\usepackage{float}
\usepackage{subfigure}
\usepackage{subcaption}
\usepackage{amsmath}
\usepackage{todonotes}
\usepackage{latexsym}
\usepackage{amssymb} 
\usepackage{multirow}
\usepackage{listings}
\usepackage{tabularx}
\usepackage{booktabs}
\usepackage{xurl}
\usepackage{seqsplit}
\usepackage{makecell}
\usepackage{ragged2e}
\usepackage{graphicx}
\usepackage{longtable} 
\usepackage{placeins}
\usepackage{supertabular}
\usepackage[square,numbers,sort&compress]{natbib}

\newcommand{\mname}{ChemMLLM}
\newcommand{\molvqgan}{Mol-VQGAN}
\newcommand{\imgcaption}{img2caption}
\newcommand{\imgproperty}{img2property}
\newcommand{\imgsmiles}{img2smiles}
\newcommand{\propertyimg}{property2img}
\newcommand{\imgimg}{img2img}

\begin{document}

\pagestyle{fancy}

\rhead{}

\title{ChemMLLM: Chemical Multimodal Large Language Model}

\maketitle

\author{Qian Tan\textsuperscript{1,2,\textdagger}}
\author{Di Zhang\textsuperscript{2}}
\author{Ben Gao\textsuperscript{3,2}}
\author{Peng Xia\textsuperscript{2}}
\author{Wanhao Liu\textsuperscript{1,2}}
\author{Shufei Zhang\textsuperscript{2}}
\author{Wanli Ouyang\textsuperscript{2}}
\author{Lei Bai\textsuperscript{2}}
\author{Yuqiang Li\textsuperscript{2,*}}
\author{Tianfan Fu\textsuperscript{4,2,*}}

{\renewcommand{\thefootnote}{\fnsymbol{footnote}}
\footnotetext[2]{This work was done during the internship at Shanghai Artificial Intelligence Laboratory.}}

\begin{affiliations}
\textsuperscript{1} Department of Electronic Engineering and Information Science, University of Science and Technology of China, Hefei, Anhui, 230026\\
\textsuperscript{2} Shanghai Artificial Intelligence Laboratory, Shanghai, 200232\\
\textsuperscript{3} The Institute for Advanced Studies, Wuhan University, Wuhan, Hubei, 430072\\
\textsuperscript{4} Computer Science Department, Nanjing University, Nanjing, Jiangsu, 210023\\
\end{affiliations}

Email Address: futianfan@gmail.com

\keywords{AI for Science, Large Language Model, AIDD, Multimodal LLM, molecule learning}

\justifying

\begin{abstract}

Recent years have seen rapid progress in multimodal large language models (MLLMs) in the field of chemistry. 
However, chemical MLLMs that can handle cross-modal understanding and generation remain underexplored. 
To fill this gap, we propose \mname, a unified chemical multimodal large language model for molecule understanding and generation. 
In this work, we design five types of multimodal tasks across text, molecular SMILES strings and images, and curate the datasets. We benchmark \mname\ against a range of general leading MLLMs, Chemical LLMs and specialized models on these tasks. Experimental results show that \mname\ achieves superior performance among
general-purpose MLLMs and close performance to specialized models across all evaluated tasks. Our work extends the capabilities of chemical multimodal large language models to the realm of image generation, demonstrating the feasibility of unifying multiple cross-modal chemical tasks within a single foundation model and enabling more intuitive, visual human–AI interaction.

\end{abstract}

\section{Introduction}

Recent advances in artificial intelligence (AI) have significantly accelerated progress in chemical research. A growing number of AI methods and models have been developed for diverse chemical tasks, including molecular property prediction, de novo molecule generation, and reaction outcome forecasting. In the chemistry area, many chemical tasks are inherently multimodal, encompassing textual descriptions, structured formats like SMILES\footnote{A SMILES (Simplified Molecular Input Line Entry System) string is a compact, text-based representation of a molecule's structure that encodes its atomic composition and connectivity in a linear format.}~\cite{weininger1988smiles}, and molecular images.

More recently, multimodal frameworks have been adapted to chemical domains, unifying traditionally disparate tasks—such as property prediction, optical chemical structure recognition (OCSR), and molecular design within single model architectures~\cite{MolScribe,liu2023multi,edwards2022translation,seidl2023enhancing,zeng2022deep,luo2023molfm,cao2023instructmol,luo2024learning,zhang2024unimot,pei2023biot5,pei2024biot5+,gao2022kg,huang2024foundation,gharizadeh2024hgtdr,li2025chemvlm,wang2025image}. These approaches generally fall into three categories: (1) Task-specialized systems that push the frontier on a single objective, such as OCSR with MolScribe~\cite{MolScribe}, property prediction with Chemprop~\cite{heid2023chemprop} and KG-based pipelines for drug repurposing~\cite{gharizadeh2024hgtdr}; (2) Representation-centric models whose primary contribution is learning transferable molecular (or cross-modal) embeddings—spanning SMILES/graph foundation models and molecule–text/knowledge alignment~\cite{liu2023multi,seidl2023enhancing,zeng2022deep,luo2023molfm,wang2023automated,luo2024learning,chenhierarchical,pei2024biot5+}; and (3) Foundation models that aim at multi-task unification, providing a single interactive interface across modalities and tasks~\cite{zhao2025developing,zhao2024chemdfm,li2025chemvlm,zhang2024chemllm,zhang2024unimot,cao2023instructmol}.

\begin{figure}[t] 
    \centering
    \includegraphics[width=18cm]{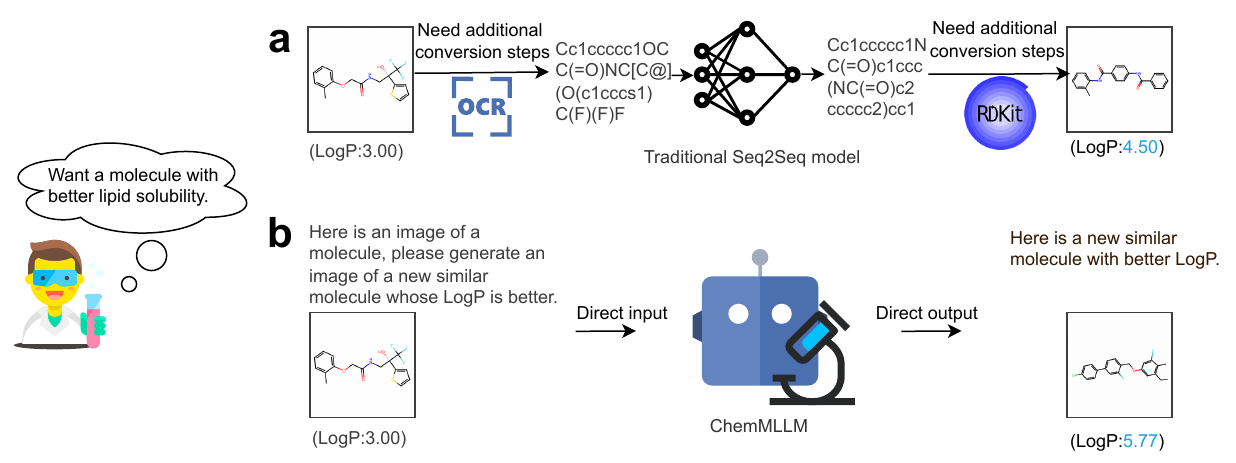}
    \caption{Schematic comparison between a conventional sequence-to-sequence (Seq2Seq)~\cite{cho2014learning} pipeline and our unified chemical large language model. a) To obtain a molecule with improved lipid solubility, conventional workflows first require recognizing the SMILES representation from a molecular image; after a Seq2Seq model predicts the SMILES, a computational chemistry toolkit (e.g., RDKit) is used to render the SMILES back into an image for visualization. b) In contrast, our unified framework (ChemMLLM) accepts the molecular image directly and produces the visualization of the generated molecule end-to-end, eliminating reliance on external conversion tools.}
    \label{fig:motivation}
\end{figure}

Despite these advancements, most existing chemical AI systems remain fundamentally limited by their task-specific designs and unidirectional information flow. They are typically capable of interpreting molecular structures from text or images but lack the bidirectional transformation and interactive ability that characterizes human expertise in chemistry. For example, while a classical Seq2Seq~\cite{cho2014learning} model may generate a SMILES string with improved properties for molecule optimization task, it cannot produce interpretable visualizations or engage in interactive dialogue about structural modifications, limiting its broader applicability. Although recent studies have proposed unified frameworks spanning multiple tasks across modalities, mainly based on multimodal large language models (MLLMs), to mitigate this issue~\cite{zhao2024chemdfm,li2025chemvlm}, most current chemical MLLMs primarily emphasize understanding. For tasks like retrieval, property prediction and optical chemical structure recognition, current chemical MLLMs mainly treat images as inputs and focus on understanding information in the image. The ability to generate chemical visuals, which reflects an essential part of how chemists communicate and interpret molecular structures~\cite{kosenkov2021computer}, remains largely absent in the literature. As shown in Fig~\ref{fig:motivation}, when utilizing models for molecular design, chemists are often restricted to symbolic outputs such as SMILES strings. To interpret these results, they must interrupt their workflow to render the structures using external tools. This dependency creates a fragmented interaction loop, significantly reducing the efficiency of the iterative design process. In contrast, direct visual generation would eliminate this bottleneck, allowing chemists to assess structures intuitively without leaving the model interface, thereby facilitating a seamless and efficient Human-Computer Interaction (HCI) paradigm. Therefore, an integrated Chemical foundation model that supports both multimodal understanding and generation for chemistry remains lacking.

\begin{figure*}[t]
\centering
\includegraphics[width=18cm]{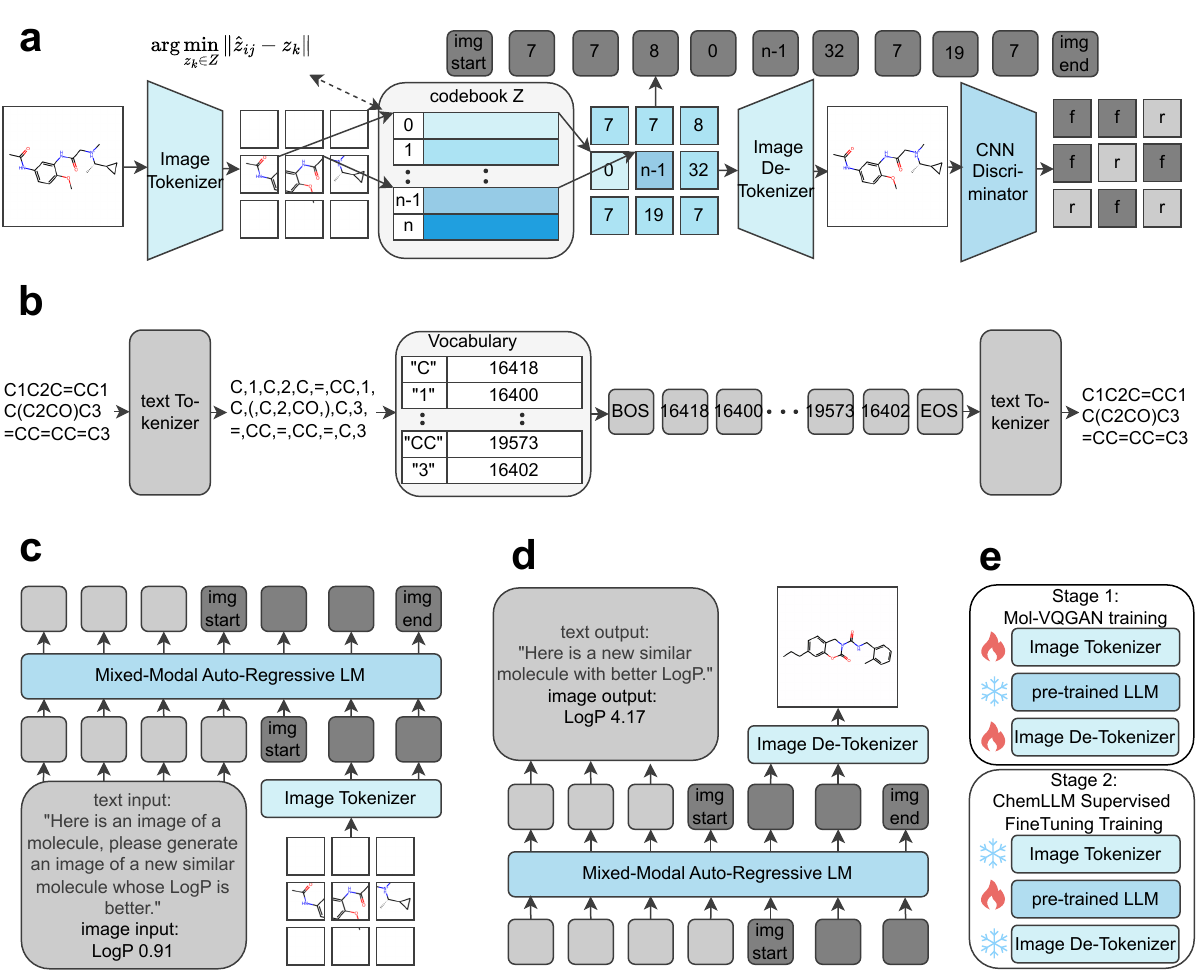}
\caption{Overall architecture of \mname.  a) Image tokenizer and de-tokenizer. The image tokenizer employs
CNN to extract spatial feature maps, where each nz-dimensional spatial code is quantized into a discrete latent code via vector quantization (VQ). The resulting codebook indices serve as the final image tokens. Image de-tokenizer uses CNN to reconstruct image from discrete feature map. Then, a patch-based discriminator predicts whether the patch is fake (f) or real (r). b) SMILES tokenizer and de-tokenizer. SMILES tokenization is consistent with text, and is mapped into a token sequence via text tokenizer.
c) \mname\ training process.
d) \mname\ inference process. 
e) two-stage training paradigm.}
\label{fig:method} 
\end{figure*}

\begin{figure}[!htbp]
\centering
\includegraphics[width=16cm]{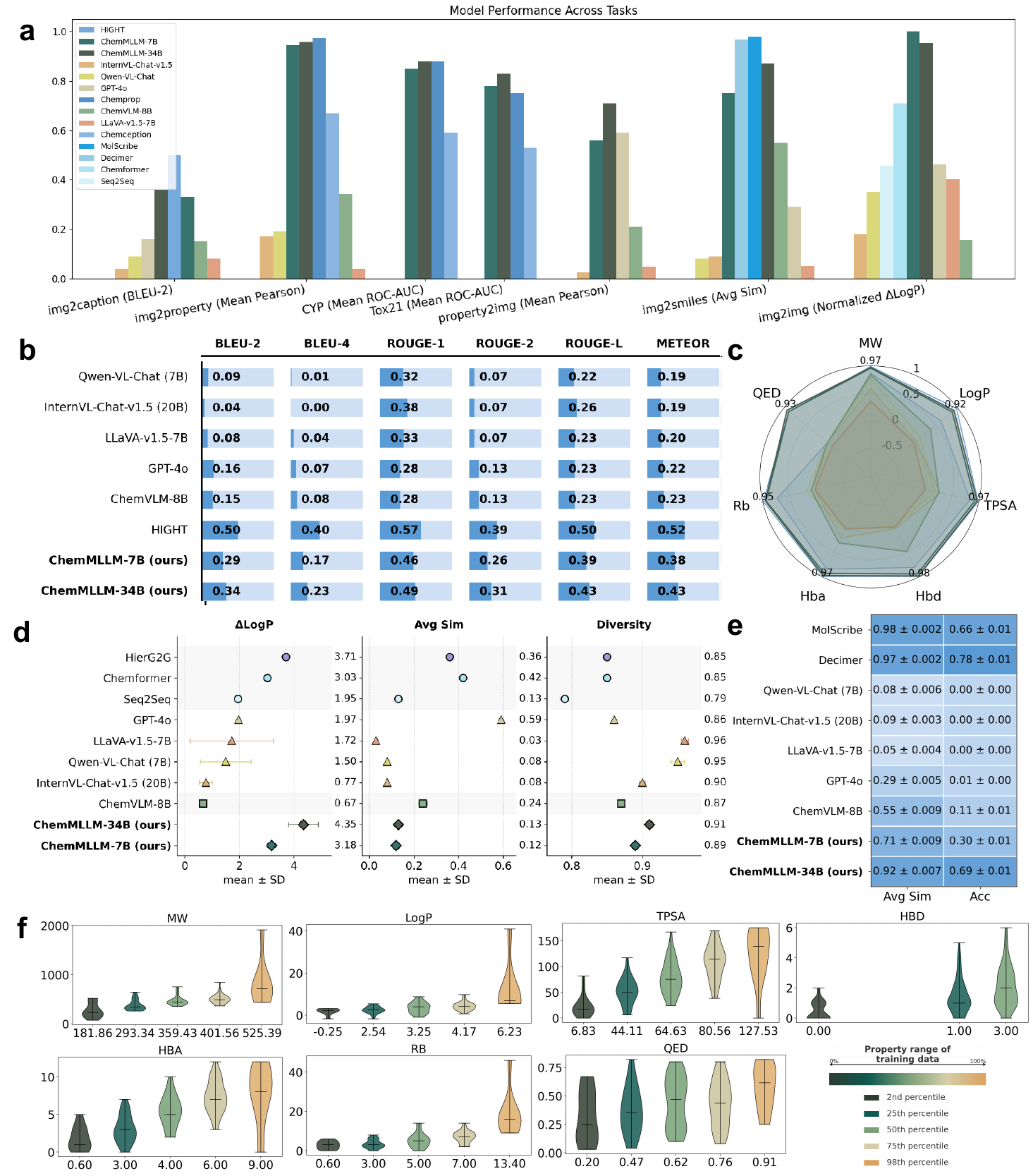}
\caption{Experimental results of five types of tasks. \mname\ achieves superior performance among
general-purpose MLLMs and close performance to specialized models across all evaluated tasks. a) Overall performance of \mname\-7B and \mname\-34B and baselines on five types of tasks. Mean Pearson means the mean value of Pearson correlation of seven properties; Avg Sim means Tanimoto similarities; Normalized $\Delta$LogP means normalized (i.e., divided by the maximum value) Increased LogP. b) Experimental results of Molecule image captioning task. c)  Experimental results of Molecule image property prediction task. The values in the radar chart represent the Pearson correlation between the predicted properties and the ground truth.d)  Experimental results of Image-to-SMILES conversion task. "Avg Sim" means
Tanimoto similarities, and “Acc” means Accuracy.  e)  Experimental results of Molecule image optimization task. $\Delta$LogP means Increased LogP. f) Experimental results of Controllable multi-objective molecule image design task at five test points based on the property distribution of its
evaluation data.}
\label{fig:main_re} 
\end{figure}

The main challenges in building such a model lies in the following three aspects: First, vision-based chemistry tasks and datasets remain underexplored. Existing chemical datasets mainly represent molecules in SMILES format. Secondly, molecule images have their own uniqueness. Unlike natural images, molecule images are sparse, containing large areas of empty space, and are composed strictly of straight lines. General MLLMs often generate unclear and distorted molecules (Fig~\ref{fig:ablation}c). Finally, how to select an effective framework for chemical MLLM to seamlessly fuse diverse modalities, including discrete SMILES string and text, as well as continuous molecule images, can be challenging. 

To address these issues, we propose \mname, a chemical multimodal large language model that understands and generates molecules in a unified framework. Specifically, to handle three challenges above, we identify five multimodal chemistry tasks with three modalities (text, SMILES, image), which contain both generation and comprehension tasks; we finetune Vector Quantized Generative Adversarial Network (VQGAN) for molecule image to bridge the gap between molecule images and natural images; we introduce the “Image Tokenizer-LLM-Image De-tokenizer” architecture into multimodal chemical tasks to fuse different modalities in early stages and enable models to generate images directly. Also, we design a two-stage training strategy and prove its effectiveness empirically. We benchmark the performance of different models on our proposed five tasks and \mname\ achieves competitive performance. The success of \mname\ provided concrete guidance for exploring multimodal large language models (MLLMs) in chemistry and promises to have wide-ranging implications for AI for science.

\section{Results and Discussion}

\begin{figure}[t]
\centering
\includegraphics[width=\linewidth]{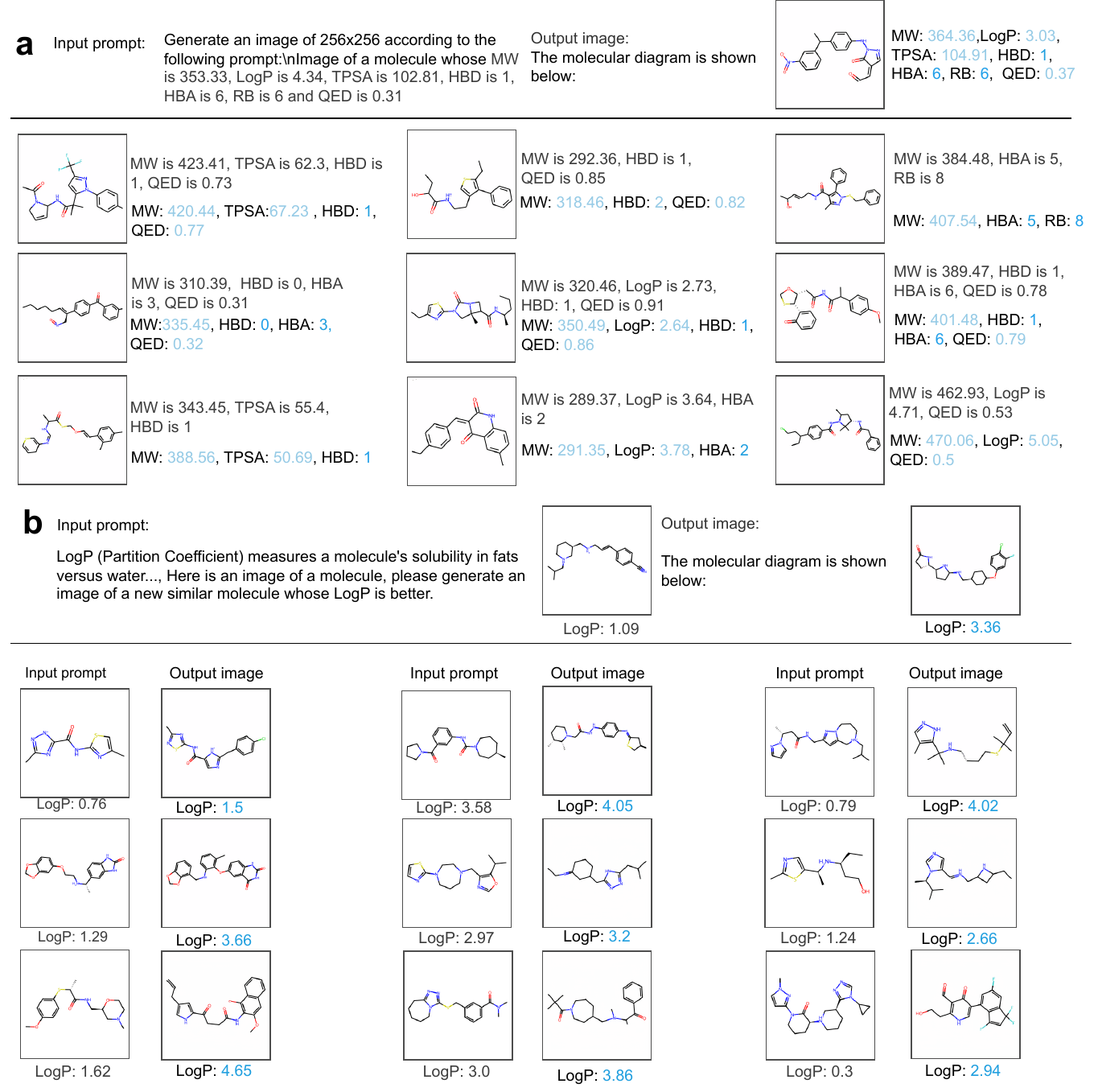  }
\caption{Examples from \mname-34B for molecule image generation tasks. a) Examples for Controllable multi-objective molecule image design task. The grey font indicates the desired properties, while the blue font represents the properties of the molecule images generated by the \mname. b) Examples for Molecule image optimization task. The left image is the input image with its LogP in grey fonts, while the right image is generated by the \mname\ with its LogP in blue fonts.}
\label{fig:case} 
\end{figure}

We design five types of vision-based chemistry research
tasks to train and evaluate our \mname\ as follows: 1) Molecule image captioning. 2) Molecule image property prediction. 3) Image-to-SMILES conversion. 4) Controllable multi-objective molecule image design. 5) Molecule image optimization. As shown in Fig~\ref{fig:main_re}a, our \mname\ chieves achieves superior performance among
general-purpose MLLMs and close performance to specialized models across all evaluated tasks. Detailed evaluation descriptions and metrics definition for each task can be seen in Supporting Information Section 4.

\subsection{Molecule Image Captioning}
Molecule image captioning is an image-to-text task, where the models are expected to generate a caption concerning the source, functionality, structure feature and usage for each molecule image. This image-to-caption task requires models to translate molecule images into natural language descriptions, which is a process mirroring how chemists annotate experimental data.

We use BLEU-2/4~\cite{papineni2002bleu}, ROUGE-1/2/L~\cite{lin2004rouge}, and METEOR~\cite{banerjee2005meteor} to evaluate the quality of generated captions against reference texts. 

We compare our model with various multimodal LLMs (MLLMs), including Qwen-VL-Chat~\cite{bai2023qwenvl}, InternVL-Chat-v1.5~\cite{chen2023internvl}, LLaVA-v1.5-7B~\cite{liu2023visual}, GPT-4o~\cite{openai2024gpt4o}, ChemVLM-8B~\cite{li2025chemvlm} and HIGHT~\cite{chenhierarchical}.  As shown in Fig~\ref{fig:main_re}b, our model exhibits strong performance on this task, outperforming most competing MLLM models excluding HIGHT on all six metrics. An example is shown in Fig~\ref{fig:ablation}d. Our model generates captions that closely match the ground truth, while Qwen-VL-Chat includes fewer semantically informative details. Given the highly template-based nature of the ChEBI-20 test set, we performed a de-templating operation by removing recurring phrases (e.g., 'the molecule is...', 'it derives from...', and 'the major microspecies at pH...') from both the ground truth references and model outputs. The re-evaluated metrics (see Supporting Information Section 2.8) indicate that our model maintains competitive performance even after eliminating these formulaic patterns.

To better demonstrate the fidelity of our model, we conducted a blind expert review for this task. Detailed results can be found in Section 11.2 of the Supporting Information.

\subsection{Molecule Image Property Prediction}

Molecule image property prediction is an image-to-text task, where models are expected to generate the value of seven different important properties for each molecule image, including molecule weight (MW), Partition Coefficient (P) of a solute between octanol and water (LogP), Topological Polar Surface Area (TPSA), Hydrogen Bond Donor (Hbd), Hydrogen Bond Acceptor (Hba), Rotatable Bond (Rb), and Quantitative Estimate of Drug-likeness (QED). This image-to-property prediction task evaluates a model’s ability to infer key chemical properties directly from 2D molecular images, enabling researchers to extract actionable insights from molecular images without specialized software, which could accelerate high-throughput screening in drug/material design~\cite{lu2021cot}.  

We extract seven molecular properties from LLM-generated outputs and evaluate the accuracy using Mean Squared Error (MSE). For this task, we do not compare with GPT-4o because it cannot predict properties from images directly, and we compare with task-specific models, including Chemception~\cite{goh2017chemception} and Chemprop~\cite{heid2023chemprop}. As shown in Fig~\ref{fig:main_re}c, our model consistently outperforms MLLMs across all seven molecular properties, yielding the highest Pearson correlation coefficients alongside the lowest MSE and MAE values. While \mname’s performance is behind specialized models like Chemprop, it offers greater versatility in handling a wider range of tasks. An example is shown in Fig~\ref{fig:ablation}e. Among the properties predicted, our model has 6 accurate values and 1 close values, while Qwen-VL-Chat has 0 close values and 7 inaccurate values.

Furthermore, we extended our evaluation to critical ADMET-related tasks using the Tox21 and CYP benchmarks. In this setup, we formulated the challenge as a multimodal understanding task by converting SMILES entries into 2D molecular images. The model is required to analyze the input image and output binary probability scores for the active class, and performance is evaluated using the Area Under the Receiver Operating Characteristic Curve (ROC-AUC). We excluded general-purpose MLLMs from this specific comparison, as they failed to yield discriminative predictions, collapsing to a trivial solution of predicting the positive class for nearly all samples. Strikingly, \mname~exhibits exceptional performance compared to domain-specific baselines. Our model surpasses the specialized graph-based model Chemprop~\cite{heid2023chemprop} on the Tox21 benchmark and achieves competitive performance with Chemprop on the CYP dataset. This result suggests that our unified framework effectively captures complex structural motifs associated with bioactivity.

\subsection{Image-to-SMILES Conversion}

Image-to-SMILES conversion is a fundamental chemistry task, where models are expected to recognize the SMILES in each molecular image. The image-to-SMILES translation task challenges models to convert 2D molecular images into SMILES strings, requiring precise recognition of atoms, bonds, rings, and stereochemistry. 

We extract SMILES strings from the LLM outputs. Then, we use accuracy that measures exact matching (based on canonical SMILES) and Tanimoto similarity that measures similarities between generated and ground truth SMILES.

In this task, we compare our model with MLLMs and task-specific models including MolScribe~\cite{MolScribe} and Decimer~\cite{rajan2020decimer}. 
The evaluation results are shown in Fig~\ref{fig:main_re}d. \mname\ achieves the best performance in both Tanimoto similarity and Accuracy metrics among all MLLMs. Despite exhibiting lower performance than MolScribe and Decimer, our model is capable of handling a broader range of downstream tasks beyond Image-to-SMILES Conversion. For Tanimoto similarity, \mname\-34B (0.92) surpasses domain-specific model ChemVLM (0.55) by 67.27\%. An example is shown in Fig~\ref{fig:ablation}f. Our model recognizes SMILES from images successfully, while GPT-4o predicts wrong SMILES with low Tanimoto similarity.

It is remarkable that ChemMLLM-34B exceeds the accuracy of the specialized MolScribe model. We further analyze the failure case of MolScribe and find that 99.64\% (277/278) of MolScribe's failures are due to incorrect stereochemistry (e.g., R/S or E/Z errors) despite correct connectivity. Since these stereochemical nuances are decisive for drug efficacy and safety, their accurate recognition is non-negotiable. These findings attest to the strength of ChemMLLM and confirm that our benchmark presents a genuine challenge, pushing the boundaries for even the most advanced domain-specific tools.

To better demonstrate the fidelity of our model, we conducted a blind expert review for this task. Detailed results can be found in Section 11.1 of the Supporting Information.

\subsection{Controllable Multi-objective Molecule Image Design}

Controllable multi-objective molecule image design is the inverse problem of molecule image property prediction and is a text-to-image task, where models are expected to generate the image of a molecule conditioned on target properties. 
It is the core of molecule design~\cite{du2022molgensurvey}. 
The challenge lies in simultaneously optimizing multiple property constraints while maintaining chemical validity. 

During our evaluation, following SketchMol~\cite{wang2025image}, generated molecule images are first converted to SMILES via MolScribe~\cite{MolScribe}, from which properties are computed. Then, the calculated properties are evaluated using MSE, MAE, and Pearson correlation against ground truth properties. Each model is run five times, and the best result is reported.

Since the GPT-4o API does not perform as well on this task as its web interface, and other MLLMs lack the capability to generate images, we treat this task as a purely text-based problem when evaluating other MLLMs. Specifically, our model is asked to generate molecular images, while other MLLMs are tasked with directly generating the corresponding SMILES strings in text form. Furthermore, we exclude Qwen-VL-Chat~\cite{bai2023qwenvl} from comparison on this task, as it fails to generate valid SMILES strings on all test samples. The evaluation results are shown in Fig~\ref{fig:main_re}f. Our model achieves top-2 or better performance on all evaluation metrics. Several examples are shown in Fig~\ref{fig:case}a. Our model generates the molecule images with desired properties directly. 

\subsection{Molecule Image Optimization}

Molecule image optimization is an image-to-image task, where models take a molecular structure with less desirable molecular properties (\textit{e.g.}, LogP) as input and generate a similar molecular structure with more desirable properties while preserving desired chemical properties. It imitates the process of lead optimization, a fundamental problem in drug discovery~\cite{huang2021therapeutics,fu2020core}. 

We use Increased LogP, which means the difference between the LogP of the generated and source molecules, as well as average tanimoto simlarity, molecular diversity and novelty to measure the optimized molecule. We utilize TDC’s evaluators: Diversity is computed as the average pairwise Tanimoto distance between ECFP fingerprints within generated molecules, and Novelty is the fraction of generated molecules not present in the training set \cite{benhenda2017chemgan,polykovskiy2018moses,huang2021tdc,rogers2010ecfp}. Other settings are the same as Controllable multi-objective molecule image design task. 

Since GPT-4o’s API underperforms its web interface on this task and other MLLMs lack image-generation capability, we evaluate them as image-to-text models (generating SMILES from images). For GPT-4o, which cannot produce optimized SMILES directly from images, we adopt a text-to-text approach, providing input SMILES as text instead. We also compare \mname\ with task-specific models including sequence-to-sequence~\cite{cho2014learning}, HierG2G~\cite{jin2020hierarchical} and Chemformer~\cite{irwin2022chemformer}. As shown in Fig~\ref{fig:main_re}d, \mname\-34B achieves the highest increase in LogP, outperforms GPT-4o by 120.81\%. Several examples are shown in Fig~\ref{fig:case}b. Our model generates molecule images with higher LogP directly.

\subsection{Direct Visual Generation versus Symbolic Rendering}
We also discuss the trade-offs between direct image generation and the traditional SMILES-render pipeline. External rendering ensures deterministic and standardized outputs but remains susceptible to error propagation from the intermediate text generation (e.g., invalid SMILES syntax). Conversely, our direct generation method circumvents this dependency, allowing for an end-to-end generation of visual objectives. This approach is particularly advantageous for capturing specific visual styles and maintaining coherence in the image domain, although it requires the model to implicitly learn chemical validity constraints. Ultimately, we view these two paradigms as complementary: external rendering excels in strict standardization, while direct generation unlocks new potential for visually-oriented human-AI interaction.

\subsection{Impact of Molecular Size and Resolution Scalability}
We quantitatively analyzed the relationship between molecular weight (MW) and visual generation fidelity (measured by LPIPS) to understand the limitations of the fixed $256 \times 256$ resolution. Our empirical results demonstrate that for typical drug-like molecules (100-1000 Da), the generation quality exhibits minimal variation with respect to molecular size, indicating that the current resolution offers sufficient information density for standard organic compounds. However, for macromolecules in the 1000--2000 Da range, we observe a slight degradation in fidelity, confirming that fixed pixel density imposes a physical bottleneck for larger and more complex structures, the detailed results can be found in Supporting Information Section 10. 
Despite this, statistical analysis of standard drug benchmarks (e.g., ZINC and ChEMBL) reveals that the vast majority of pharmaceutical candidates fall well below this threshold, suggesting the current setup is robust for primary molecule applications. 

Regarding the scalability, to enable the model to adapt to higher resolutions, such as $512 \times 512$, no additional architectural modifications are required. The process simply entails retraining the VQGAN on higher-resolution data, followed by retraining the LLM. The primary distinction between resolutions is that larger images will be encoded into a greater number of discrete tokens.

\subsection{The Necessity and Complementary Role of the Visual Modality}
Despite the fact that molecules are intrinsically defined by their topology and 3D geometric conformations, the integration of a 2D visual modality addresses critical gaps in Human-Computer Interaction (HCI) and real-world data accessibility that symbolic or 3D geometric models alone cannot fulfill. From the perspective of interaction, 2D depictions constitute the primary cognitive interface for chemists. Unlike 1D SMILES strings, which act as non-intuitive symbolic intermediates requiring external rendering tools for verification, or 3D structures that often necessitate specialized software for manipulation, generated 2D images offer immediate and lightweight visual feedback. Direct image generation eliminates the fragmented workflow of translating symbolic outputs into visuals, thereby reducing cognitive load and significantly accelerating the iterative design loop between the researcher and the AI. Furthermore, the visual modality bridges the disconnect between the physical and digital worlds, unlocking data sources that remain inaccessible to models lacking image processing capabilities. In educational and collaborative settings, non-standardized inputs such as hand-drawn sketches or blackboard notes are prevalent yet represent a ``blind spot'' for standard graph models. Endowing the model with visual understanding allows users to query these inputs directly without manual translation to SMILES or graph. Moreover, a vast amount of chemical knowledge exists solely as bitmap images within historical literature, patents, and laboratory notebooks, lacking corresponding machine-readable text or 3D coordinates. An image-capable foundation model is uniquely positioned to mine and interpret this unstructured data. Thus, compared with 3D geometric information and 1D SMILES, the 2D image modality serves as a vital complement, enhancing the model's usability, interpretability, and coverage of real-world chemical data.

\subsection{Interpretability and Visual Grounding Analysis}
\label{sec:discussion_interpretability}
To investigate the underlying mechanisms driving our model's predictions, we conducted an interpretability study focusing on the visual attention dynamics during the Image-to-SMILES generation process.

We first examined the faithfulness of the learned attention maps via a counterfactual perturbation analysis. Specifically, we aggregated the cross-modal attention from generated tokens to the input image grid to identify highly attended regions. We then performed a top-$k$ occlusion test, masking the top $k\%$ of these regions and measuring the impact on generation performance. We find the results show a distinct degradation: masking only a small fraction of highly attended regions causes a substantial drop in accuracy, and this degradation amplifies as $k$ increases (see Supporting Information Section 12). These counterfactual results confirm that the model's predictions are not driven by spurious background noise, but are causally dependent on specific, visually grounded structural evidence.

To further elucidate what chemical features are being learned, we visualized the cross-attention distributions. As illustrated in Fig~\ref{fig:ablation}g, we observed two distinct patterns: (a) the global mean cross-attention broadly covers the entire molecular skeleton, indicating a holistic perception of the topology; and (b) the token-specific attention for distinct functional groups (e.g., amide, nitro, sulfonyl, and carboxylic acid) is sharply localized to their corresponding visual substructures. This accurate alignment indicates that the model establishes a robust structure--semantics grounding between 2D molecular depictions and textual chemical concepts, effectively moving beyond simple pattern matching to capture the underlying chemical essence.

\subsection{Performance Analysis of Direct SMILES Generation and Image Generation}
To rigorously compare the performance of image-based versus SMILES-based molecular generation, we compared our model against a baseline trained exclusively on SMILES sequences for Molecule Image Property Prediction and Controllable Multi-objective Molecule Image Design tasks. We observed that the SMILES-only baseline achieves superior numerical performance (see Supporting Information Section 2.7). This is anticipated, given the discrete and highly structured nature of SMILES strings compared to the complexity of high-dimensional image generation.

However, we emphasize that the primary contribution of our work is not to surpass text-based benchmarks numerically, but to extend the capabilities of language models to the visual domain. A SMILES-only approach fundamentally lacks the ability to process or produce molecular depictions, which limits its utility in real-world scenarios where chemical knowledge is curated visually (e.g., in patents, literature, and laboratory notebooks). By enabling direct image understanding and generation, our model provides a complementary interface that bridges the gap between textual data and the ubiquitous visual language of chemistry, offering a multimodal capability that text-only models cannot replicate.

\subsection{Ablation Study} 
\paragraph{Impact of \molvqgan\ and LLM SFT.}

We conduct an ablation study on the molecule image property prediction task to evaluate the individual contributions of the domain-specific \molvqgan\ and the LLM Supervised Fine-Tuning (SFT). We utilize a subset of 200 samples from the \propertyimg\ task and perform a 5-shot evaluation. 

The full model, equipped with both the domain-adapted \molvqgan\ and LLM SFT, achieves the most robust performance across the majority of molecular properties. Specifically, it attains high Pearson correlation coefficients for MW (0.71), TPSA (0.71), Hba (0.66), and Rb (0.62), demonstrating that the synergy between the visual tokenizer and the fine-tuned language model effectively captures semantic fidelity and chemical structure.

When the domain-specific \molvqgan\ is replaced (i.e., w/o VQGAN), we observe a sharp decline in performance for structural metrics, despite retaining LLM SFT. For instance, the correlation for MW drops significantly to 0.24, and the model fails to capture polarity-related features, as evidenced by near-zero or negative correlations for TPSA (-0.02) and Hba (-0.06). Although the QED score remains comparable (0.36), the collapse in other structural metrics underscores the critical role of \molvqgan\ in encoding spatial chemical information that the LLM alone cannot infer.

Furthermore, removing LLM SFT while keeping the VQGAN results in a complete failure of the model. The performance drops to near-random levels across all metrics (e.g., -0.05 for MW and -0.01 for TPSA). This indicates that without chemical-specific instruction tuning, the LLM lacks the necessary alignment to interpret visual tokens and reason about molecular properties. In conclusion, both the specialized visual tokenizer and the supervised fine-tuning are indispensable for accurate chemical multimodal understanding. Detailed results can be found in Supporting Information Section 2.6. 

\begin{figure}[t]
\centering
\includegraphics[width=\textwidth]{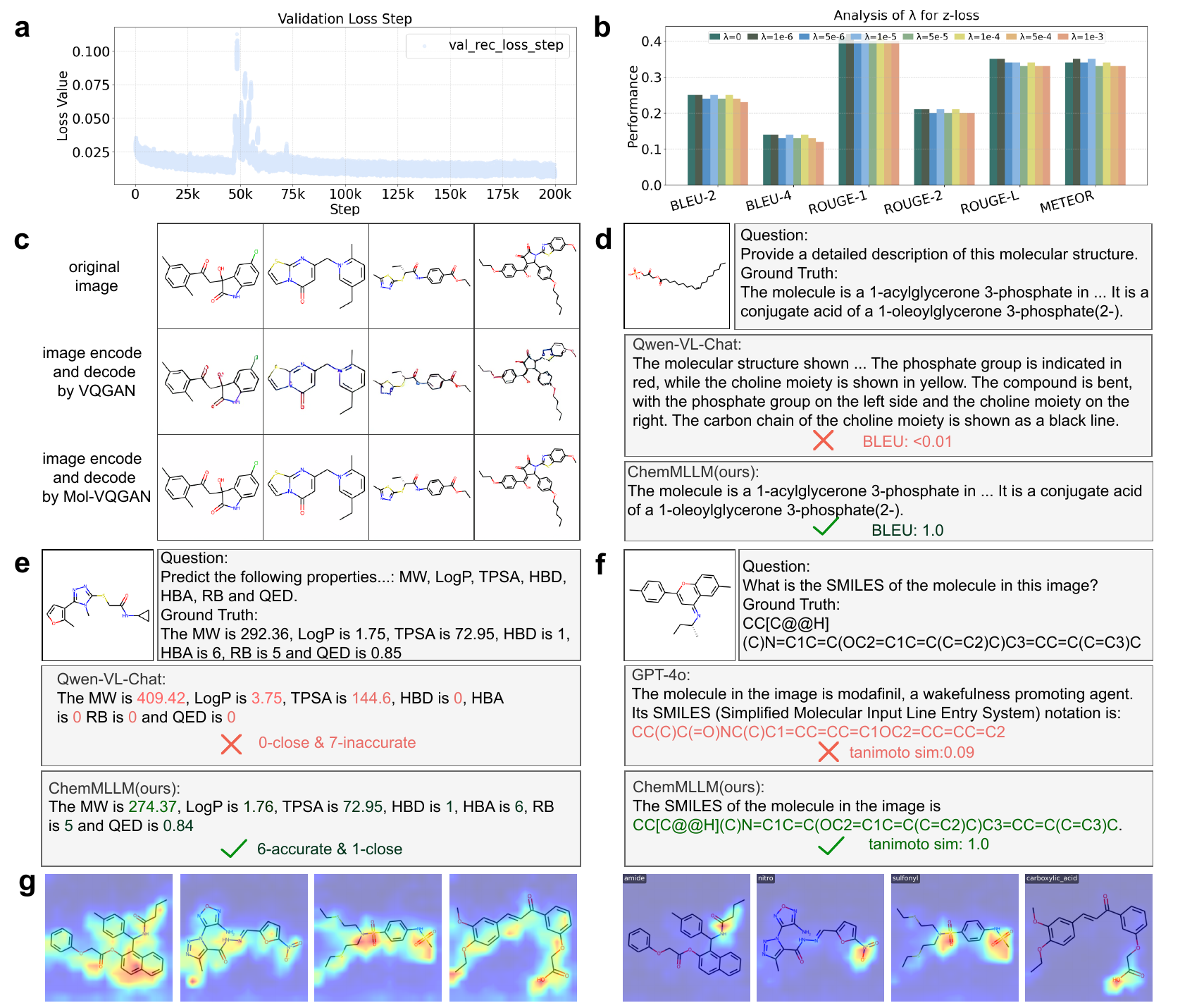}
\caption{More visualized experiment results. a) Training curve from start to the best checkpoint, we apply GAN loss after training in E,G, and Z for a period of steps for stability. As shown in the validation loss curve, the observed spike marks the onset of adversarial training, where the discriminator loss is introduced, transitioning the model from a pure reconstruction objective to the full VQGAN objective. The loss converges naturally without the need for any special handling. b) Results on Molecule image captioning task with different $\lambda$ c) Examples of origin VQGAN and \molvqgan. Origin images are shown in the first row, results for origin VQGAN are shown in second row and results for \molvqgan\ is shown in the third row. The origin VQGAN can not encode and decode molecule image clearly and accurately, resulting in distorted atoms and
bonds that are hard to distinguish, while our well-trained \molvqgan can encode and decode molecule images with clear atoms and bonds, which is almost the same as the origin molecule images. d) An example on molecule image captioning task, comparison between Qwen-VL-Chat and our \mname-34B. e)  comparison of answers on molecule image property prediction task on Qwen-VL-Chat and our \mname-34B. Accurate answers are highlighted in bottle-green, close answers are highlighted in light-green and inaccurate answers are highlighted in red. f) A comparison of GPT-4o’s and \mname-34B’s answers on Image-to-SMILES conversion task. g) Cross-attention heatmaps from model-generated tokens to image tokens. Left: Cross-attention over all generated output tokens, showing the global alignment between the generated description and the molecular image.
Right: Cross-attention restricted to functional-group tokens (amide, nitro, sulfonyl, and carboxylic acid), highlighting localized image regions associated with each predicted functional group.}
\label{fig:ablation} 
\end{figure}

\paragraph{Analysis of $\lambda$}
To systematically investigate the influence of the hyperparameter in Eq.~\eqref{eq:llm_loss}, we conduct experiments by evaluating eight distinct values of 
$\lambda$ on the molecule image captioning task. As illustrated in Fig~\ref{fig:ablation}b, our experimental results demonstrate that the variation in $\lambda$ exhibits only a marginal impact on model performance. In alignment with the methodology adopted in Lumina-mGPT~\cite{liu2024lumina}, we consequently fix $\lambda$ at $1e-5$ for all LLM training procedures.

\section{Conclusion}

This paper has proposed \mname, a chemical multimodal large language model that handles molecule comprehension and generation across text, SMILES string, and molecule image seamlessly. Our \mname\ is composed of a \molvqgan\ tokenizer and a large language model (LLM). The \molvqgan\ discretizes and reconstructs images into codebook tokens, which allows the LLM to treat images as sequences analogous to text, and LLM is responsible for generating text and meaningful image tokens. The combination of \molvqgan\ and \mname\ enables early, seamless multimodal fusion.

To address the gap that VQGAN trained on general images can not faithfully encode and decode molecule depictions, we retrain the VQGAN on a large corpus of molecule images so that it acquires domain-specific image representation capacity. For better multimodal alignment, we adopt a two-stage training scheme. At the first stage, we train the VQGAN. Then, we freeze the VQGAN and use it as the image tokenizer for the LLM. The first stage enables the VQGAN to learn chemically faithful representations of molecule images, while the second stage injects this visual molecule knowledge into the LLM to improve downstream performance. Also, to mitigate the challenge of scarce data for multimodal chemical tasks, we design five types of cross-modal chemistry tasks and curate corresponding datasets, providing a valuable resource for multimodal AI in chemistry. 

We benchmark the performance of different models on our proposed five tasks. Across thorough empirical studies, results demonstrate that \mname\ outperforms state-of-the-art MLLMs and achieves results comparable to specialized models, highlighting its strong performance and potential for real-world drug and material discovery. More importantly, \mname\ overcome the limitation of prior chemical multimodal foundation models which could not natively generate molecule images and instead produced SMILES followed by external rendering. This capability makes it possible for a single multimodal language model to unify diverse multimodal tasks in chemistry, opening up new avenues for the application of multimodal large language models in the field of chemistry.

\section{Methods}
\begin{figure}[t]
\centering
\includegraphics[width = \textwidth]{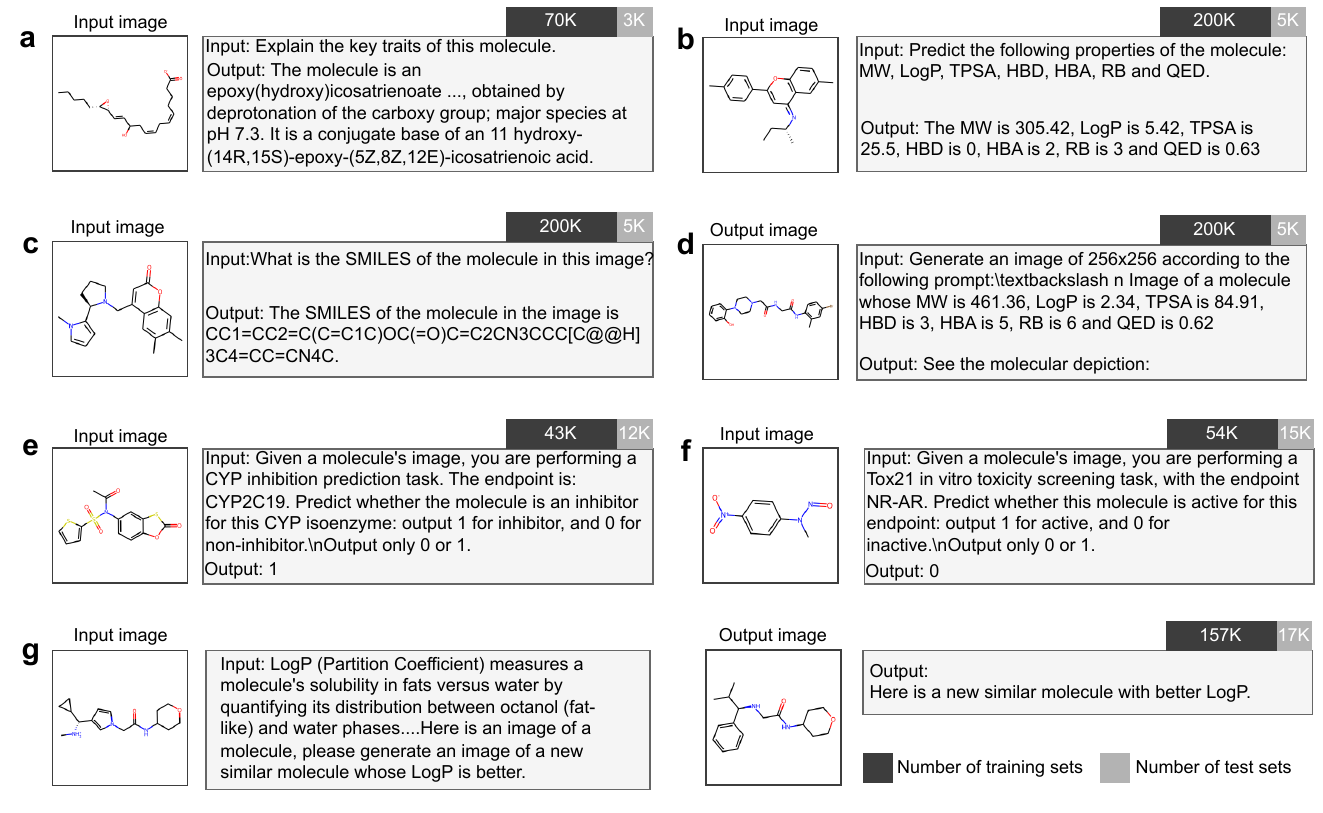}
\caption{Example data and statistic information of training and test set of our designed five tasks. a) Example and statistic for molecule image captioning task. b) Example and statistic for molecule image property prediction task. c) Example and statistic for Image-to-SMILES conversion task. d) Example and statistic for controllable multi-objective molecule image design task. e) Example and statistic for CYP task. f) Example and statistic for Tox21 task. g) Example and statistic for molecule image optimization task.}
\label{fig:example} 
\end{figure}
\subsection{\mname\ Framework and  Loss Function}
As shown in Fig~\ref{fig:method}, to unify the understanding and generation for chemical tasks, our framework uses an image tokenizer to transfer images into discrete tokens, aligning with texts and molecule's SMILES strings at the token level, known as ``Image Tokenizer-LLM-Image De-tokenizer'' architecture~\cite{team2024chameleon}.
We first discuss the molecule image tokenizer, and then describe how to combine images, texts and SMILES strings in Multimodal LLM, followed by the training strategy. 

\paragraph{Mol-VQGAN for Molecule Image Generation.}
\label{sec:vqgan}

Following Chameleon~\cite{team2024chameleon}, to enable multimodal information alignment in the early stage, images need to be discretized into token sequences similar to text. To achieve this, we train Vector Quantized Generative Adversarial Network (VQGAN)~\cite{esser2021taming} on molecule images, known as \molvqgan. \molvqgan\ compresses images into a discrete latent space, which aligns well with the sequential nature of text modeling in large language models. 
Specifically, \molvqgan\ uses Vector Quantized Variational Auto-Encoder (VQVAE)~\cite{van2017neural} as the generator and patch-based discriminator~\cite{isola2017image}. VQVAE compresses images into discrete spaces and reconstructs images from that space. \molvqgan\ adds a discriminator and perceptual loss to VQVAE to keep good perceptual quality and improve the performance. 
Formally, \molvqgan\ takes an image $x \in \mathbb{R}^{H \times W \times 3}$ ($H/W$ are the height/width of the input image), and transfers it into discrete representations $z_\textbf{q}$ by encoding $\hat{z} \in \mathbb{R}^{h \times w \times n_z} = E(x)$ ($h/w/n_z$ are the height/width/channels of the feature map), and finding the closest codebook entry for each spatial code $\hat{z}_{ij}\in \mathbb{R}^{n_z}$ (also known as vector quantization process, denoted $\textbf{q}(\cdot)$): 
\begin{equation}
\begin{aligned}
z_q =  \textbf{q}(\hat{z})
    = \big( \arg \min_{z_k \in Z} \| \hat{z}_{ij} - z_k \| \big) \in \mathbb{R}^{h \times w \times n_z},
\end{aligned}
\end{equation} 
where Z is codebook, a set of learnable vectors, and $z_k \in Z ={\{z_i\}}_{i=1}^n \subset \mathbb{R}^{n_z}$. 
Then, \molvqgan\ reconstructs image $\hat{x} \in \mathbb{R}^{H \times W \times 3}$ from $z_\textbf{q}$:
\begin{equation}
\hat{x}=G(z_q)=G(\textbf{q}(E(x))).
\end{equation}
The discretization and reconstruction process is illustrated in Fig~\ref{fig:method}a. 
The Mol-VQGAN's objective function is: 
\begin{equation}
\begin{aligned}
& \min_{E,G,Z} \max_{D} \big[\mathcal{L}_{vqvae}(E,G,Z) + \lambda_1 \mathcal{L}_{perceptual}(E,G,Z)
+ \lambda_2 \mathcal{L}_{GAN}(\{E,G,Z\},D) \big],  
\end{aligned}
\end{equation}
where the first term $\mathcal{L}_{vqvae}(E,G,Z) = ||x-G(\hat{z}-sg(\hat{z}-z_q))||_2^2+||sg[\hat{z}]-z_q||^2_{2} +||sg[z_q]-\hat{z}||^2_{2}$ is VQVAE loss, where $\mathcal{L}_{rec} = ||x-G(\hat{z}-sg(\hat{z}-z_q))||_2^2$ is reconstruction loss. Since the quantization process is non-differentiable, a stop-gradient operation $sg[\cdot]$ is used so that the forward pass operates on quantized vectors, whereas the backward pass leverages continuous vectors for gradient computation~\cite{van2017neural}; the second term $\mathcal{L}_{perceptual}(E,G,Z) 
= ||P(x)-P(G(\hat{z}-sg(\hat{z}-z_q)))||_2^2$ is perceptual loss, $P$ denotes a perceptual model like Learned Perceptual Image Patch Similarity (LPIPS)~\cite{zhang2018unreasonable}, which is used to extract the high-level semantic features; the third term $\mathcal{L}_{GAN}(\{E,G,Z\},D)=\log D(x)+\log(1-D(\hat{z}-sg(\hat{z}-z_q)))$ is GAN loss, $D$ is a patch-based discriminator~\cite{isola2017image} aiming to differentiate original and reconstructed images. $\lambda_1$ is a hyperparameter and $\lambda_2$ is an adaptive weight computed dynamically to balance the weight of $\mathcal{L}_{GAN}$ and stabilize training, following VQGAN~\cite{esser2021taming}.

\paragraph{\mname.}
\label{sec:chemmllm}

Chemical molecules are typically encoded in the format of Simplified Molecular Input Line Entry System (SMILES)~\cite{weininger1988smiles}, a compact ASCII string, serving as a distinct modality in computational chemistry~\cite{anderson1987smiles}. In \mname, SMILES tokenization is the same as the text. Specifically, SMILES is mapped into a token sequence via Chameleon~\cite{team2024chameleon} text tokenizer trained based on Byte Pair Encoding (BPE) algorithm~\cite{sennrich2015neural}. The process is shown in Fig~\ref{fig:method}b.

\mname\ uses a well-trained \molvqgan\ as image tokenizer to align image and text at token level, unifying the training and inference for image and text by maximizing the standard next-token prediction cross-entropy loss: 
\begin{equation}
\begin{aligned}
 \mathcal{L}_{}
     =  \sum_{i=1}^L \log p_\theta(s_i|s_1,...,s_{i-1}) 
    + \lambda \sum_{k}(\log \sum_{j=1}^V \exp(z_{k,j}))^2,
\end{aligned} 
\label{eq:llm_loss}
\end{equation}
where $\lambda$ is a hyper-parameter; in the first term, $L$ is the length of total sequence tokens; $s_i \in S=\{S_I,S_T\}$. $S_I$ is the image tokens sequence tokenized by the image tokenizer, and $S_T$ is the text/SMILES token sequence tokenized by the text tokenizer. 
The second term is z-loss, a regularization term to mitigate the problem of logit shift in the final softmax and stabilize training~\cite{chowdhery2023palm}, where $z_{k,j}$ denotes the logit in the last layer, $V$ denotes the size of the vocabulary.

The image tokenizer takes images of 256×256 resolution as input. Simultaneously, text and SMILES information pass through the text tokenizer to be converted to text tokens. The text, SMILES and image tokens are then concatenated to form a unified token sequence to feed into the LLM during training and inference.

\subsection{Two-Stage Training Methods}
\label{sec:training}

As observed in Fig~\ref{fig:method}e, \mname\ is trained in two complementary phases: 1) \molvqgan\ training. 2) Multimodal sequence learning. The first phase specializes the image tokenizer to molecule images while the second phase freezes the image tokenizer and trains a decoder-only LLM to operate over a unified token stream of images, SMILES, and text.

\paragraph{\molvqgan\ Training.}
The original Chameleon VQGAN is only trained on the natural image dataset and can not discretize and reconstruct molecule images well. So, the first stage focuses on improving VQGAN's performance in encoding and decoding molecule images. 
Concretely, we use the well-trained VQGAN~\cite{esser2021taming} (trained on natural images) as the initialization and then fine-tune it on molecule image datasets. For training code, we use the official implementation code of the original VQGAN~\cite{esser2021taming}. We sample 1,000,000 molecule images from PubChem synthesized by RDKit~\cite{landrum2006rdkit} and combine them with all images synthesized in 5 downstream tasks as training dataset for \molvqgan. All parameters of Encoder, Decoder and codebook of VQGAN are trained on 8 NVIDIA A800×80G GPUs for two epochs. The batch size is set to 16 and the base learning rate is set to 4.5e-06. We use Adam~\cite{kingma2014adam} as optimizer and $\mathcal{L}_{rec}+\lambda_1 \mathcal{L}_{perceptual}$ as monitor to validate the model and save the best checkpoint. The validation curve from start to the best checkpoint is shown in Fig~\ref{fig:ablation}a.

\paragraph{Multimodal Sequence Learning.}
In the second stage, we freeze the \molvqgan\ and only finetune the language model on five downstream tasks. For training code, we utilize Lumina-mGPT~\cite{liu2024lumina}. We trained models at two scales: 7B and 34B. For base model, we use Chameleon-7B as the base model for \mname-7B and Lumina-mGPT-34B-512 for \mname-34B. The weights related to the image tokens in the last layer will first be initialized as zero during finetuning. The training hyper-parameters primarily follow Lumina-mGPT~\cite{liu2024lumina}. We train \mname-7B on 8 NVIDIA H200×140G GPUs and \mname-34B on 16 NVIDIA H200×140G GPUs. Other training details are provided in Section 6 of the Supporting Information. 

\subsection{Data Curation Details}

\label{app:data_curation_details}
\paragraph{Data Collection.}
In this work, we utilize ChEBI-20~\cite{edwards2022translation}, 
Mol-Instructions~\cite{fang2023mol}, PubChem~\cite{kim2021pubchem}, HierG2G~\cite{jin2020hierarchical} and TDC~\cite{jin2020hierarchical} as data source. All the data used in this paper is publicly available. For ChEBI-20~\cite{edwards2022translation}, we download training, test and val data in .txt format in the data subfolder, the original data contains 26,407 train data, 3,300 test data and 3,301 val data. Each data entry contains a SMILES-caption pair; for
Mol-Instructions~\cite{fang2023mol}, we download the corresponding json files from Molecule-oriented\_Instructions.zip, the original data contains SELFIES-caption pairs with instructions. We first transfer SELFIES into SMILES using RDKit~\cite{landrum2006rdkit}, then we filter captions shorter than 150 words to screen clearer descriptions and finally get 43,392 samples; for PubChem~\cite{kim2021pubchem}, we download all the structure data files (SDF) in the Compound subfolder up to January 13, 2025. Then we use Chem.SDMolSupplier() function in RDKit~\cite{landrum2006rdkit} to parse the original SDF and get 9,691,316 compounds with three fields each related to SMILES, \textit{i.e.}, "PUBCHEM\_SMILES" (the SMILES for compounds), "PUBCHEM\_OPENEYE\_CAN\_SMILES" (canonical SMILES generated using the OpenEye tool~\cite{OpenEye2023}), and "PUBCHEM\_OPENEYE\_ISO\_SMILES" (the isomeric smiles generated using the OpenEye tool). After that, we utilize the set operation to remove the duplicate SMILES and finally get 14,874,837 SMILES from the raw dataset. for HierG2G~\cite{jin2020hierarchical}, we download train\_pairs.txt and test.txt from logp04 and logp06 subfolder. The original data contains 157,673 training data and 17,520 test data. All data entries contain paired SMILES representations, arranged in ascending order of LogP values. For the Tox21 task, we downloaded all 12 subset datasets from the Therapeutics Data Commons (TDC) website, totaling 70,156 data points. Similarly, for the CYP task, we retrieved data for all 5 subcategories from TDC, comprising a total of 56,515 samples. The raw data consists of a SMILES string and a binary label (0 or 1) indicating the activity status (active or inactive). Then we utilize the Draw.MolToImage() function in RDKit~\cite{landrum2006rdkit} to transfer SMILES into 256$\times$256 images for all datasets.

\paragraph{Data Synthesis.}
All data employed in this work are synthetically constructed based on existing datasets, using different processing pipelines depending on whether natural language descriptions were originally available. For datasets that already contain textual descriptions, such as ChEBI-20~\cite{edwards2022translation} and Mol-Instruct~\cite{fang2023mol}, we simply converted the SMILES representations into images to create multimodal versions. In cases where the task is well-defined but lacks natural language descriptions, such as the Tox21 and CYP benchmarks, we generated images from SMILES and additionally designed natural language templates, which were then populated with the structured labels (e.g., 0/1) to form the textual component. For custom tasks—specifically Molecule Image Property Prediction, Image-to-SMILES Conversion, Controllable Multi-objective Molecule Image Design, and Molecule Image Optimization—we first extracted SMILES from PubChem and calculated their physicochemical properties using RDKit~\cite{landrum2006rdkit}. These calculated attributes were then filled into predefined templates to generate descriptions~\cite{li2025chemvlm}, paired with their corresponding image renderings. For Molecule image captioning task, the dataset is sourced from ChEBI-20~\cite{edwards2022translation} and Mol-Instructions~\cite{fang2023mol}. The examples for this task can be seen in Fig~\ref{fig:example}a; for Molecule image property prediction task, the dataset is sourced from PubChem~\cite{kim2021pubchem}. We choose 7 important properties as the prediction objective, \textit{i.e.}, MW, LogP, TPSA, HBD, HBA, RB and QED. Then use RDKit to calculate the 7 properties for each sampled SMILES. Then we apply natural language templates to integrate properties into natural language to form the image-property answer pairs. The templates for this task can be seen in Fig~\ref{fig:example}b; for Tox21 and CYP tasks, the dataset is sourced from TDC~\cite{huang2021tdc}, the templates can be seen in Fig~\ref{fig:example}e and Fig~\ref{fig:example}f; for Image-to-SMILES conversion task, the dataset and the construction steps are the same as the Molecule image property prediction task, the difference is that this task only applies templates for SMILES to construct image-SMILES answer pairs. The templates for Image-to-SMILES conversion task can be seen in Fig~\ref{fig:example}c; Controllable multi-objective molecule image design task is the inverse of Molecule image property prediction task. By swapping the question and answer of the the dataset, we construct property prompt-image pairs for Controllable multi-objective molecule image design task. The templates for this task can be seen in Fig~\ref{fig:example}d; for Molecule image optimization task, the dataset is sourced from HierG2G~\cite{jin2020hierarchical}. The templates for this task can be seen in Fig~\ref{fig:example}g.

\begin{table}[h]
\centering
\small
\resizebox{1\linewidth}{!}{
\begin{tabular}{ccccccc}
\toprule[1pt]
{Task} & Input & Output & {Source} & \# train & \# test(total) & \# test(used)  \\
\midrule
\makecell[c]{\imgcaption} & \makecell[c]{image\\+text} & text & \makecell[c]{ChEBI-20~\cite{edwards2022translation}\\ Mol-Instructions~\cite{fang2023mol}} & 70K&3K &3k\\ 
\hline 
\makecell[c]{\imgproperty} & \makecell[c]{image\\+text} & text & PubChem~\cite{kim2021pubchem} & 200K&5K & 1k\\ 
\hline
\makecell[c]{Tox21} & \makecell[c]{image\\+text} & text & TDC~\cite{huang2021tdc} & 54K & 15K & 15k \\
\hline 
\makecell[c]{CYP} & \makecell[c]{image\\+text} & text & TDC~\cite{huang2021tdc} & 43K & 12K & 12k\\
\hline 
\makecell[c]{\imgsmiles} & \makecell[c]{image\\+text} & SMILES & PubChem~\cite{kim2021pubchem} & 200K&5K & 1k\\
\hline 
\makecell[c]{Controllable multi-objective \\ molecule image design } & text & image & PubChem~\cite{kim2021pubchem} & 200K & 5K & 200\\
\hline 
\makecell[c]{\imgimg } & \makecell[c]{image\\+text} & image & HierG2G~\cite{jin2020hierarchical} & 157K & 17K & 200\\
\bottomrule
\end{tabular}
}
\vspace{-2mm}
\caption{Tasks and datasets. }
\label{table:data_split}
\end{table}

\paragraph{Data Split.}
Detailed specifications of data sources and split sizes are provided in Table~\ref{table:data_split}. For Molecule Image Captioning, we combined the ChEBI-20 training set with a filtered description subset from Mol-Instruct for training, using the ChEBI-20 test set for evaluation. For the Property Prediction, Image-to-SMILES, and Controllable Design tasks, we randomly sampled 205,000 entries from PubChem and the random seed is set to 42, split with a 40:1 train-test ratio. For Tox21 task, we perform a scaffold-based split into training and test sets utilized the official scaffold split implementation provided by TDC. And for CYP tasks, we create the training/test split with a fixed random seed of 42 using the official get\_split method provided by TDC. For Molecule Image Optimization, the train and test dataset was derived from the HierG2G~\cite{jin2020hierarchical} training set using a 9:1 split. Regarding the evaluation detail, for established benchmarks such as CYP, Tox21, and ChEBI-20, we performed evaluations on the full test sets to ensure consistency and fair comparison with prior studies. In contrast, for the custom tasks introduced in this work, we evaluated a randomly sampled subset of the test data to facilitate efficient evaluation. The exact number of samples actually used for evaluation is reported in the \# test (used) column in Table~\ref{table:data_split}.

\medskip
\textbf{Acknowledgements} \par 
This work was supported by the New Generation Artificial Intelligence—National Science and Technology Major Project (Grant No. 2025ZD0121802). The AI-driven experiments, simulations, and model training were performed on the robotic AI-Scientist platform at the Chinese Academy of Sciences.

\medskip
\textbf{Author Contributions} \par
Q.T. designed the task and carried out the experiment. T.F., Y.L., W.O., and L.B. supervised the investigations. Q.T. and W.L. performed the experiments. P.X. and D.Z. analyzed the data. Q.T., B.G., and T.F. wrote the paper. All authors have read and agreed to the published version of the manuscript.

\medskip
\textbf{Resource availability} \par 
All data used in this paper are publicly available and can be accessed at the following link after the paper is published. \url{https://github.com/bbsbz/ChemMLLM.}

The code for training \molvqgan\ are available at: \url{https://github.com/CompVis/taming-transformers.} The code for training \mname\ are available at: \url{https://github.com/Alpha-VLLM/Lumina-mGPT.} The ChEBI-20 dataset is available
at: \url{https://github.com/cnedwards/text2mol.} The Mol-Instructions dataset is
available at: \url{https://huggingface.co/datasets/zjunlp/Mol-Instructions.} The PubChem dataset
is available at: \url{https://misuse.ncbi.nlm.nih.gov.}
The TDC dataset is available at: \url{https://github.com/wengong-jin/iclr19-graph2graph.}

\medskip
\textbf{Declaration of interests} \par
The authors declare no competing interests.

\medskip
\textbf{Declaration of generative AI and AI-assisted technologies} \par
During the preparation of this work, the authors used GPT-5 (OpenAI) and Gemini (Google) in order to improve the readability and language of the manuscript, as well as to assist in the creation and refinement of figures. After using this tool/service, the authors reviewed and edited the content as needed and take full responsibility for the content of the publication.

%

\bibliographystyle{unsrtnat}   
\bibliography{references}




\end{document}